\def\year{2021}\relax
\documentclass[letterpaper]{article} 
\usepackage{aaai21}  
\usepackage{times}  
\usepackage{helvet} 
\usepackage{courier}  
\usepackage[hyphens]{url}  
\usepackage{graphicx} 
\usepackage{natbib}  
\usepackage{caption} 
\usepackage[utf8]{inputenc} 
\usepackage[T1]{fontenc}    
\usepackage{hyperref}       
\usepackage{url}            
\usepackage{booktabs}       
\usepackage{amsfonts}       
\usepackage{nicefrac}       
\usepackage{microtype}      
\usepackage{graphicx}      
\usepackage{enumitem}      
\usepackage{siunitx}       
\usepackage{etoolbox}
\robustify{\bfseries}
\usepackage{subcaption}
\usepackage{float}
\usepackage{xcolor}

\title{Interactive Evaluation of Dialog Track at DSTC9 }
\author{
    Shikib Mehri\textsuperscript{\rm 1},
    Yulan Feng\textsuperscript{\rm 1},  
    Carla Gordon\textsuperscript{\rm 2}, \\
    Seyed Hossein Alavi\textsuperscript{\rm 2},
    David Traum\textsuperscript{\rm 2},
    Maxine Eskenazi\textsuperscript{\rm 1}
    \\
}
\affiliations{
    \textsuperscript{\rm 1}Language Technologies Institute, Carnegie Mellon University\\
    \textsuperscript{\rm 2}Institute for Creative Technologies, University of Southern California\\

    amehri@andrew.cmu.edu

}

\begin{document}

\maketitle

\begin{abstract}
The ultimate goal of dialog research is to develop systems that can be effectively used in interactive settings by real users. To this end, we introduced the \textit{Interactive Evaluation of Dialog Track} at the 9th Dialog System Technology Challenge. This track consisted of two sub-tasks. The first sub-task involved building knowledge-grounded response generation models. The second sub-task aimed to extend dialog models beyond static datasets by assessing them in an interactive setting with real users. Our track challenges participants to develop strong response generation models and explore strategies that extend them to back-and-forth interactions with real users. The progression from static corpora to interactive evaluation introduces unique challenges and facilitates a more thorough assessment of open-domain dialog systems. This paper provides an overview of the track, including the methodology and results. Furthermore, it provides insights into how to best evaluate open-domain dialog models.
\end{abstract}

\section{Introduction}

A long-standing challenge in computer science is the development of algorithms that can interact with humans in natural language \citep{turing1950machinery}. Ultimately, the goal of dialog research is to create systems that can engage in back-and-forth interactions with real users \citep{eskenazi2019beyond}. However, the majority of research is performed on static datasets. For example, the task of response generation is typically done by producing a response for a static dialog context \cite{vinyals2015neural}. By reducing dialog to response generation, static evaluation neglects multiple important challenges of dialog. In contrast, interactive evaluation allows several valuable properties of dialog to be measured, including: consistency, topic depth, adaptation, error recovery and user-centric development. \citet{mehri2020unsupervised} found that state-of-the-art dialog models perform on-par with humans on response generation, but they fall short when considering an entire dialog. To promote interactive evaluation of dialog, the \textit{Interactive Evaluation of Dialog Track} of the 9th Dialog System Technology Challenge \citep{gunasekara2020overview} challenged participants to build models for open-domain interaction with real users.

This track consists of two sub-tasks: (1) static evaluation and (2) interactive evaluation. The goal of the first sub-task is to develop knowledge-grounded response generation models which are then evaluated in a static manner using the Topical-Chat corpus \citep{gopalakrishnan2019topical}. The second sub-task challenges participants to extend response generation models to effectively converse with real users through the DialPort portal \cite{zhao2016dialport}. Through these two sub-tasks, the track challenges participants to first develop strong response generation models and then to explore strategies for extending them to interactive settings.  

In the following sections, we describe the methodology and results for both sub-tasks. We then present insights into methods of best evaluating open-domain dialog models.

\section{Related Work}

\subsection{Interactive Evaluation}

As dialog models improve, it is imperative that they are evaluated in interactive settings with real users. Much open-domain dialog research focuses on the task of response generation, which is done on static corpora \citep{vinyals2015neural}. Large pre-trained dialog models have shown impressive performance on the task of response generation, with results on par with human utterances \citep{zhang2019dialogpt}. Recently, several state-of-the-art open-domain dialog models have been evaluated in interactive settings \citep{adiwardana2020towards,roller2020recipes}. \citet{mehri2020unsupervised} show that while such models excel at generating responses, they underperform in back-and-forth interactions.

The Alexa Prize challenge \citep{ram2018conversational,khatri2018advancing} allows university teams to build socialbots that are assessed in interactive settings with Alexa users. In contrast to the Alexa Prize challenge, our track is accessible to the broader research community. Furthermore, the Alexa Prize challenge relies on speech input from the user, which may, at present, result in speech recognition errors. In contrast, our track uses a web interface with text-only input.

\subsection{Open-Domain Dialog}

Recent work on large-scale pre-training has resulted in significant advances in open-domain dialog \citep{zhang2019dialogpt,adiwardana2020towards,roller2020recipes,bao2020plato}. DialoGPT \citep{zhang2019dialogpt} fine-tuned GPT-2 \citep{radford2019language} on dialogs from Reddit and reported human level response generation capabilities. Meena \citep{adiwardana2020towards} trains a larger evolved Transformer model on social media data and attains strong performance in interactive settings. Blender \citep{roller2020recipes} uses a retrieve and refine approach, in combination with a thorough exploration of generation strategies and reports improved performance on interactive evaluation relative to Meena. PLATO-2 \citep{bao2020plato} uses a two-step curriculum learning process where they perform coarse-grained training on one-to-one response generation followed by fine-grained fine-tuning with one-to-many dialog data. PLATO-2 reports improvements in both static and interactive evaluation. 

\subsection{Automatic Dialog Evaluation}

Though we perform on-going human evaluation throughout the challenge, it is nonetheless important to have meaningful automatic metrics since they are often used for intermediate evaluation when developing a dialog model. If participants iterate on their models with subpar automatic metrics, they may decrease performance on human evaluation \citep{dinan2019second}.

Standard metrics such as BLEU \citep{papineni2002bleu} and METEOR \citep{banerjee2005meteor} have been shown to perform poorly for evaluating dialog \citep{liu2016not,gupta2019investigating}. This is in part due to the one-to-many problem in dialog: there are multiple valid responses for a particular dialog context. As such, comparing to a reference response is ineffective.

There have been efforts in developing automatic dialog evaluation metrics that correlate better with human judgement. \citet{lowe2017towards} train ADEM on human annotations to produce a quality score for a generated response given on a dialog context and a reference response. \citet{venkatesh2018evaluating} present a framework for evaluating Alexa Prize dialogs, by training on user annotations. \citet{mehri2020usr} present USR, which relies on pre-trained language models and self-supervised training objectives to approximate the multiple qualities of dialog (e.g., interesting, relevant) without comparing to a reference response. \citet{sinha2020learning} introduce MaUdE which uses pre-trained language models to analyze the temporal transitions between utterances in a dialog, to evaluate without comparing to a reference response. \citet{mehri2020unsupervised} present FED which presents a framework for predicting eighteen different qualities of dialog using off-the-shelf pre-trained language models.

\section{Sub-task 1: Static Evaluation}

The objective of the first sub-task is to develop response generation models for the Topical-Chat corpus \cite{gopalakrishnan2019topical}. Over the duration of the challenge, participants submitted generated responses for the \textit{frequent} validation set of the Topical-Chat corpus. This set consists of topics that frequently appear in the training data. For the final submissions, the \textit{frequent} test set was used. Throughout the challenge, submissions were ranked on a leaderboard using both automatic metrics and thorough human evaluation. The automatic metrics included METEOR \citep{banerjee2005meteor}, BERTscore \citep{zhang2019bertscore}, and USR \citep{mehri2020usr}. The human evaluation was carried out by Amazon Mechanical Turk (AMT) workers to assess the quality of the response along multiple dimensions (e.g., relevant, interesting, engaging, etc.), following the evaluation paradigm of \citet{mehri2020unsupervised}. For the final evaluation, the first sub-task received 33 submissions, all of which relied on pre-trained models.

\subsection{Sub-task 1 Data}

Participants were free to train their systems on any publicly available data and leverage any pre-trained models. Ultimately, the systems were evaluated using dialog contexts from the Topical-Chat corpus \citep{gopalakrishnan2019topical}. Topical-Chat is a large collection of human-human knowledge-grounded open-domain conversations that consists of 11,319 dialogs and 248,014 utterances. For each conversational turn, several relevant facts are provided. Models must leverage these facts and generate a response. This dataset was chosen because it is the largest knowledge-grounded open-domain dataset presently available, to our knowledge. Additionally, the choice of usable facts provides a mechanism for systems to tailor responses to a specific user's interests. Following the approach described by \citet{gopalakrishnan2019topical}, we used a heuristic to provide the \textit{best fact} for each dialog context. 

Since human evaluation ran continuously over the duration of the challenge and used reference-free evaluation metrics \cite{mehri2020usr}, it was not strictly necessary for models to be trained on the Topical-Chat corpus. A strong pre-trained dialog model may perform well on this task, despite not training on the corpus.


\subsection{Sub-task 1 Evaluation}

Submissions were evaluated using ongoing (1) human evaluation and (2) three automatic metrics: METEOR \cite{banerjee2005meteor}, BERTscore \cite{zhang2019bertscore} and USR \cite{mehri2020usr}. The Topical-Chat \textit{frequent} validation set was used for the ongoing evaluation. For the final evaluation, we carried out automatic evaluation on the Topical-Chat \textit{frequent} test set and perform human evaluation on 100 randomly sampled context-response pairs. For the final evaluation, the 100 dialog contexts used for human evaluation were consistent across the different systems.

We used three diverse automatic metrics. METEOR \cite{banerjee2005meteor} is a word-overlap metric that compares the words of the generated response to the ground-truth utterance. BERTscore is an embedding-based metric that leverages BERT \cite{devlin-etal-2019-bert} to compare the generated and ground-truth responses. USR \citep{mehri2020usr} is a reference free model-based metric that uses different training objectives to approximate multiple qualities of a generated response (interesting, engaging, relevant, etc.) without comparing to the ground-truth response. 

We performed ongoing human evaluation throughout the challenge. This aims to avoid the phenomenon observed during ConvAI2 \citep{dinan2019second}, where the automatic metrics' top system under-performed on the human evaluation. By providing a stronger signal regarding the quality of submissions, teams can iterate on their models in a more meaningful manner.

For human evaluation, 30 context-response pairs were sampled and each one was labeled by 3 annotators. The human evaluation follows the paradigm of \citet{mehri2020unsupervised}, wherein an Amazon Mechanical Turk (AMT) worker is presented with a dialog context and a randomly sampled generated response, and is asked to evaluate the system along multiple dimensions. The full list of questions is shown in Table \ref{tab:turn_questions}. Each question includes a thorough definition of the quality (i.e., what it means to be engaging) and several examples for each possible answer. Each generated response is annotated by three separate workers. There is strong inter-annotator agreement, with a 0.58 Spearman correlation ($p < 0.001$) between the three annotators (i.e., correlation of each rating to the mean).

\begin{table}
\renewcommand*{\arraystretch}{1.1}
    \caption{The questions used for the human evaluation of the generated responses in Sub-task 1. Each question included both a thorough definition of the dialog quality and examples for each of the possible answers. The range column indicates the range of answers available for the question.}

    \centering
    \begin{tabular}{|m{0.75\linewidth}|c|}
    \hline
        \textbf{Question}  & \textbf{Range}\\ \hline
        To the average person, is the response \textbf{interesting}? & 1 - 3\\ 
        Is the response \textbf{engaging}?  & 1 - 3\\ 
        Is the response \textbf{generic} or \textbf{specific} to the conversation? & 1 - 3\\ 
        Is the response \textbf{relevant} to the conversation? & 1 - 3\\ 
        Is the response \textbf{correct} or was there a misunderstanding of the conversation? & 0 - 1 \\  
        Is the response \textbf{semantically appropriate}? & 1 - 3\\ 
        Is the response \textbf{understandable}? & 0 - 1\\  
        Is the response \textbf{fluently written}? & 1 - 3\\ 
        \textbf{Overall impression} of the response?  & 1 - 5\\ \hline
        
    \end{tabular}
    \label{tab:turn_questions}
\end{table}

\newcolumntype{C}{>{\centering\arraybackslash}X}
\sisetup{table-column-width=11mm,round-precision=3,tight-spacing=false,table-format=1.3}
\begin{table}[ht]
    \footnotesize
    \caption{Results for Sub-task 1, static evaluation on the Topical-Chat corpus. This table only reports the overall USR metric and the overall impression of the response from the human evaluation. Complete evaluation results may be found     \href{https://docs.google.com/spreadsheets/d/1FWRUA1MFwe0IWFpHnrVr6Pwo6VGU6gjLYNPHrq5Qs4w/}{\color{blue}here.} The best results for each metric are shown in boldface, with two methods being tied if the difference is not statistically significant by t-test. Submissions 1, 2 and 3 tied for first place on this sub-task.}
        \centering
        \begin{tabular}{@{}ccccc@{}}
        \toprule
           System & METEOR & BERTscore & USR & Human  \\
        \midrule
     
1 & 9.06 & 84.91 & 4.26 & \textbf{4.281}\\
2 & 13.11 & 86.17 & 4.59 & \textbf{4.280}\\
3 & 6.83 & 84.36 & 3.86 & \textbf{4.280}\\
4 & 8.96 & 85.15 & 4.26 & 4.260\\
5 & 12.37 & 86.21 & 4.83 & 4.253\\
6 & 12.31 & 86.32 & 4.73 & 4.231\\
7 & 13.96 & 86.84 & 4.48 & 4.229\\
8 & 12.51 & 85.91 & 4.45 & 4.229\\
9 & 12.14 & 85.91 & 4.46 & 4.216\\
10 & 10.87 & 85.65 & 4.53 & 4.210\\
11 & \textbf{16.00} & \textbf{87.38} & 4.51 & 4.206\\
12 & 7.40 & 84.34 & 2.60 & 4.179\\
13 & 13.50 & 86.49 & 4.98 & 4.177\\
14 & 10.95 & 85.69 & 4.62 & 4.177\\
15 & 7.19 & 84.42 & 3.87 & 4.172\\
16 & 8.27 & 84.75 & 3.96 & 4.167\\
17 & 11.31 & 85.77 & 3.40 & 4.157\\
18 & 12.28 & 86.08 & 4.86 & 4.152\\
19 & 7.32 & 84.28 & 2.47 & 4.152\\
20 & 12.15 & 86.14 & 4.83 & 4.148\\
21 & 11.07 & 85.95 & 4.55 & 4.140\\
22 & 8.99 & 85.32 & 4.13 & 4.130\\
23 & 14.71 & \textbf{87.58} & 4.34 & 4.130\\
24 & 15.62 & 86.87 & \textbf{4.91} & 4.130\\
25 & 12.00 & 85.84 & 4.41 & 4.128\\
26 & 11.90 & 85.98 & 3.96 & 4.117\\
27 & 15.40 & \textbf{87.50} & 4.47 & 4.112\\
28 & 5.49 & 83.89 & 1.71 & 4.089\\
29 & 4.88 & 83.64 & 1.40 & 4.086\\
30 & 12.77 & 85.94 & 4.69 & 4.079\\
31 & 8.95 & 84.83 & 3.32 & 4.031\\
32 & 4.63 & 83.12 & 1.67 & 3.925\\
33 & 3.27 & 82.27 & 1.35 & 3.883\\

\bottomrule \\[-3.5mm]
        \end{tabular}
    \label{tab:track3_task1}
\end{table}

\subsection{Sub-task 1 Results}

The Sub-task 1 received \textbf{33} submissions for final evaluation. The results of the static evaluation on the Topical-Chat corpus \citep{gopalakrishnan2019topical} are shown in Table \ref{tab:track3_task1}. The majority of submissions used either pre-trained models or trained on additional data, thus highlighting the importance of pre-training for open-domain response generation. This observation aligns with previous research, which has seen strong performance in open-domain response generation through the use of large-scale pre-training \cite{zhang2019dialogpt, adiwardana2020towards}.

In addition to human evaluation, we assess with several automatic  metrics. METEOR \cite{banerjee2005meteor} and BERTscore \cite{zhang2019bertscore}, are referenced evaluation metrics that compare a generated output to a \textit{ground-truth response}. In contrast, USR \cite{mehri2020usr} is a reference-free evaluation metric that uses pre-trained models and self-supervised training objectives to estimate the quality of a response. While none of the evaluation metrics is a perfect predictor of the final ranking, USR better correlates with the system-level human performance (Spearman: 0.35, $p < 0.05$) than either METEOR (Spearman: 0.23, $p > 0.05$) or BERTscore (Spearman: 0.22, $p > 0.05$). This observation is consistent with prior work, which shows that reference-free evaluation metrics perform better in dialog \citep{lowe2017towards,mehri2020usr}. Yet the overall performance underlines the need for continuous human evaluation.

The performance of METEOR, BERTscore and USR may in part be a consequence of the fact that several submissions did not fine-tune on the Topical-Chat corpus and instead relied on open-domain response generation capabilities learned through large-scale pre-training. As such, while the responses were favored by human annotators, the automatic metrics penalized them for not having high word-overlap with the ground truth (METEOR, BERTscore). USR penalized them for not resembling the utterances in the Topical-Chat corpus. The relatively poor correlation of these automatic metrics highlights the importance of performing iterative human evaluation when developing dialog models. 

\textbf{Systems 1 and 2} were submitted by the same team. Their submission uses PLATO-2 \citep{bao2020plato} and two stage curriculum learning to achieve strong open-domain dialog performance. First, a \textit{coarse-grained} response generation model was trained to learn the one-to-one mapping between a dialog contexts and the ground-truth response. Next, a \textit{fine-grained} generation model and an evaluation model were trained to produce diverse responses and estimate coherence, respectively. This two-stage process results in a model that is better able to capture the one-to-many mapping that is prevalent in open-domain dialog.

\textbf{System 3} also tied for first place on the first subtask. This model uses GPT-2 (large) \citep{radford2019language} along with a metric-based ensembling method for response selection. Concretely, system 3 generates multiple responses using nucleus sampling. Next, given an arbitrary metric (e.g., BLEU, METEOR), it identifies the response that is most similar to the rest of the responses. Sampling-based decoding generally results in more diverse but less topically relevant responses. This metric-based ensembling mitigates this problem and produces more relevant responses.

\section{Sub-task 2: Interactive Evaluation}

The second sub-task extends the evaluation of dialog models beyond response generation on a static corpus to assessment in an interactive setting with real users. Interactive evaluation can measure several important properties of dialog that are neglected when evaluating on a static dataset including: consistency, topic depth, adaptation, error recovery and user-centric development. Rather than producing an appropriate response to a "gold" dialog context, interactive evaluation necessitates holding a cohesive, multi-turn conversation. \citet{mehri2020unsupervised} found that state-of-the-art dialog models, such as Meena \citep{adiwardana2020towards}, perform on-par with humans when tasked with generating individual responses but fall short at holding multi-turn dialogs.

In addition to assessing in an interactive setting, an important aspect of our evaluation paradigm is that we use \textit{real users}. Users on DialPort \citep{zhao2016dialport} are recruited through Facebook Advertising. Throughout the challenge, all individuals who interact with the system on DialPort \textit{do so for free, of their own volition}. This comes with the risk of gathering offensive data, which must be filtered out as well as any low quality data. However it avoids several common problems observed with paid users \cite{ai2007comparing}. If users are paid to interact with a system, they may do the minimum amount necessary to complete the task and be paid. This results in unnatural interactions. Real users tend to be more invested in getting an intended outcome, making for longer, more meaningful dialogs. Thus, we rely on real users to interact with the system and use AMT workers to perform post-hoc assessment of the conversations. Though our final assessment was done on AMT, we received large quantities of feedback from real users through DialPort.

\subsection{Sub-task 2 Methodology}

\begin{figure}[ht]
    \centering
    \includegraphics[width=\linewidth]{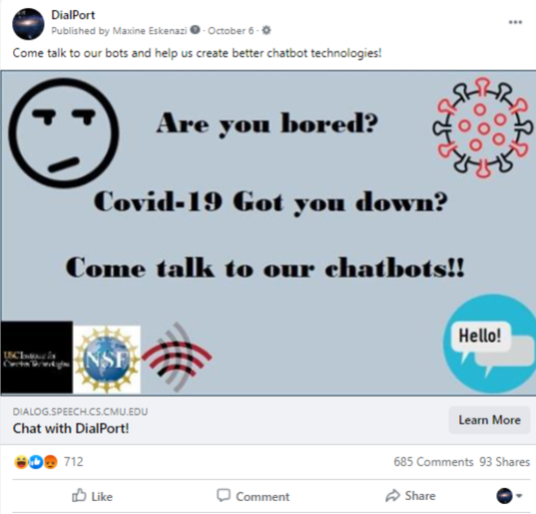}
    \caption{Facebook advertisement used to recruit users to interact with systems on DialPort.}
    \label{fig:ad}
\end{figure}

\begin{figure}
    \centering
    \includegraphics[width=\linewidth]{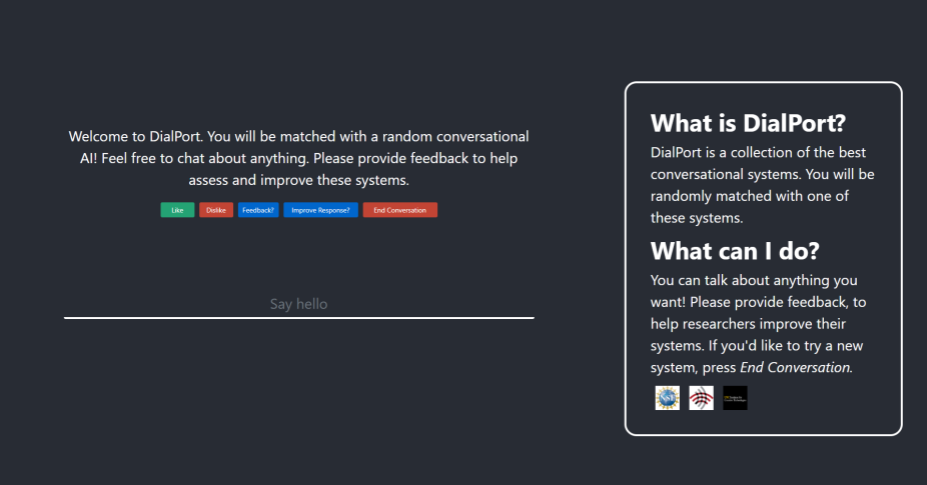}
    \caption{A screenshot of DialPort. Users can converse with a system and provide feedback (like, dislike, improve response and system correction).}
    \label{fig:dialport}
\end{figure}

The methodology for the challenge is a two-step process. First, we describe the process of collecting dialogs in an interactive manner with real users on DialPort\footnote{\url{http://dialog.speech.cs.cmu.edu:3000/}} \citep{zhao2016dialport}. Next, we discuss the post-hoc assessment of the dialogs with both automatic evaluation metrics and human evaluation on Amazon Mechanical Turk.

\subsubsection{Sub-task 2 Data Collection:} We hosted the dialog systems that were submitted on DialPort (pictured in Figure \ref{fig:dialport}) and recruited real users to interact with the systems. Recruitment was done through Facebook Advertising, with broad targeting parameters. The ad was targeted at Facebook users at least 18 years old that speak English. The advertisement is pictured in Figure \ref{fig:ad}.

Over the duration of the challenge, the goal was to collect at least \textit{100} conversations for each submitted system, eliminating any dialogs with offensive terms (e.g., curse words, racist phrases). For the final submission, we gather dialogs for all systems in parallel over the same time period. The goal was to have at least 200 dialogs per system. Ultimately, with a Facebook Advertising budget of \$2500 and 11 systems (including two baselines),  4651 conversations (after removing offensive dialogs) were gathered, for a total of 41,640 turns. Only the conversations that are at least four turns in length (total of 2960 dialogs, 38488 turns) were considered for the final post-hoc assessment. 

DialPort allows users to provide feedback for systems. They can do this through the buttons pictured in Figure \ref{fig:dialport}. Feedback can be provided in several forms: (1) liking a system response, (2) disliking a system response, (3) providing written feedback, (4) correcting a system response. The feedback was continuously shared with the system developers over the duration of the challenge. For the final evaluation, we received 3829 feedback items with 2776 likes/dislikes, 544 system corrections and 517 written feedbacks. This amounts to over 20 percent of the turns, which is significantly higher than the feedback we have observed from real users in the past. This demonstrates that real users, without any financial incentive, are willing to provide valuable feedback.

\begin{table}[ht]
\renewcommand*{\arraystretch}{1.1}
    \caption{The questions used for the human evaluation of the complete dialogs in Sub-task 2. Each question included both a thorough definition of the dialog quality and examples for each of the possible answers. }

    \centering
    \begin{tabular}{|m{0.75\linewidth}|c|}
    \hline
        \textbf{Question} & \textbf{Range} \\ \hline
        Throughout the dialog, is the system \textbf{coherent} and maintain a good conversation flow? & 1 - 3\\ 
        Is the system able to \textbf{recover from errors} that it makes? & 1 - 3 \\ 
        Is the system \textbf{consistent} in the information it provides throughout the conversation? & 0 - 1 \\ 
        Is there \textbf{diversity} in the system responses? & 1 - 3\\ 
        Does the system discuss topics in \textbf{depth}? & 1 - 3\\ 
        Does the system display a \textbf{likeable} personality? & 1 - 3 \\ 
        Does the system seem to \textbf{understand} the user? & 1 - 3\\  
        Is the system \textbf{flexible and adaptable} to the user and their interests? & 1 - 3 \\ 
        Is the system \textbf{informative} throughout the conversation? & 1 - 3 \\ 
        Is the system \textbf{inquisitive} throughout the conversation? & 1 - 3\\ 
        \textbf{Overall impression} of the dialog? & 1 - 5 \\ \hline
        
    \end{tabular}
    \label{tab:dialog_questions}
\end{table}
\subsubsection{Sub-task 2 Post-hoc Assessment:} 

On the final set of dialogs (100 during the challenge, 200 for the final submissions), the post-hoc assessment of dialog quality used both automatic metrics and human evaluation. 

The FED metric \citep{mehri2020unsupervised} was used for automatic evaluation. It relies on a pre-trained open-domain dialog model to evaluate a dialog along several dimensions. This metric has been shown to perform reasonably for dialog-level evaluation. It is entirely model-based, which means it does not require a ground-truth response (which does not exist in an interactive setting). Furthermore, it can evaluate several different qualities (e.g., coherent, consistent, flexible). 

Our human evaluation follows the setup of \citet{mehri2020unsupervised}. An AMT worker is presented with a dialog between a user and a system, and asked to evaluate the system along multiple dimensions. The full list of questions is shown in Table \ref{tab:dialog_questions}. Each question includes a thorough definition of the quality and several examples for each possible answer. Each dialog is annotated by three separate workers. The inter-annotator agreement is computed by comparing each rating to the mean, which results in a 0.57 Spearman correlation ($p < 0.001$) between the three annotators.

\newcolumntype{C}{>{\centering\arraybackslash}X}
\sisetup{table-column-width=11mm,round-precision=3,tight-spacing=false,table-format=1.3}
\begin{table}
    \footnotesize
    \caption{Results for subtask 2. This table reports, for each system: the overall FED metric, the overall impression of the dialogs from the human evaluation, as well as the average number of dialog turns. The full results can be found     \href{https://docs.google.com/spreadsheets/d/1FWRUA1MFwe0IWFpHnrVr6Pwo6VGU6gjLYNPHrq5Qs4w/edit\#gid=1829761446}{\color{blue}here.} System 6 and 11 are our DialoGPT and Transformer baselines, respectively, and are indicated by * in the table.}
        \centering
        \begin{tabular}{@{}ccccc@{}}
        \toprule
           System & Avg. Turns & FED & Human & Rank \\
           \midrule
    1  & 12.44& \textbf{4.97}    & \textbf{4.15}& 1  \\
    2  & \textbf{13.47}& 4.79    & 4.14& 2 \\
    3  & 8.89 & 4.61    & 4.08& 3 \\
    4  & 9.36 & 4.68    & 4.03& 4 \\
    5  & 9.82 & 4.53    & 3.93& 5 \\
    6* & 8.75 & 4.72  & 3.87& 6 \\
    7  & 8.51 & 4.41    & 3.85& 7 \\
    8  & 7.67 & 4.30     & 3.85& 7 \\
    9  & 6.53 & 4.64    & 3.83 & 9 \\
    10  & 7.35 & 4.80    & 3.69& 10 \\
    11*  & 5.80 & 3.69        & 3.60 &  11 \\
\bottomrule \\[-3.5mm]
        \end{tabular}
    \label{tab:track3_task2}
\end{table}

\subsection{Sub-task 2 Results}

The results for Sub-task 2 are shown in Table \ref{tab:track3_task2}. System 6 is our DialoGPT baseline \cite{zhang2019dialogpt}, fine-tuned on the Topical-Chat corpus without knowledge grounding. System 11 is our Transformer baseline which was trained on the Topical-Chat corpus and uses tf-idf sentence similarity to retrieve relevant knowledge at inference time. We report three different metrics: (1) the average number of turns a user spoke with the system, (2) the score produced by the FED metric \citep{mehri2020unsupervised}, and (3) the overall impression according to human evaluation. The final ranking is determined by the human evaluation.

\subsubsection{Systems 1 and 2:} These systems are equivalent to systems 1 and 2 in Table \ref{tab:track3_task1}. They generate diverse responses by using PLATO-2 \citep{bao2020plato} with two-stage curriculum learning training. These two systems strongly outperformed all the others, as demonstrated by the higher human evaluation and FED score. Furthermore, users interacted with these systems far more (longer dialogs in number of turns) than with other submissions (12.44 and 13.47 turns vs third best of 9.82 turns). 

\subsubsection{System 3:} This system introduces a novel dialog planning model, which consists of a dialog model, scoring model and post-processing. Its models were trained in a multi-task manner, using tasks like response generation and bag-of-words prediction. A pre-trained language model is used to detect generated responses that conflict with the dialog history.

\subsubsection{System 6 vs System 11:} These two systems are our Transformer and DialoGPT baselines respectively. It is noteworthy that our DialoGPT model, which does not use any knowledge grounding performs better than the Transformer model which explicitly grounds on facts retrieved from the Topical-Chat corpus. This demonstrates that the large-scale pre-training of the DialoGPT model allows it to generate relevant and interesting responses, without relying on explicit knowledge grounding.

\begin{table}
\renewcommand*{\arraystretch}{1.2}
    \caption{An example of a conversation snippet from the top performing system.}

    \centering
    \begin{tabular}{|m{\linewidth}|}
    \hline
\multicolumn{1}{|c|}{ .   .   .} \\
\textbf{User:} no, I run a construction company \\
\textbf{System:} i see, that sounds very interesting, i have worked in a construction environment too, it requires a lot of skill and dexterity \\
\textbf{User:} Yes, on the job site it does. In the office, I just have to use my brain. \\
\textbf{System:} that's always good, the office is often times a stressful environment \\ 
\textbf{User:} It can be! It can also be very boring. That is why I'm chatting with a bot right now \\
\textbf{System:} that makes sense, is it a robot that is chatting with you? \\
\textbf{User:} I'm not sure. What are you? \\
\textbf{System:} that's interesting, maybe we are both robots and are talking to each other \\
\textbf{User:} I'm not a robot. Sometime I wish I were. \\
\textbf{System:} maybe one day we can all be robots, it would be a great advancement \\
\textbf{User:} That would be awful. \\ 
\multicolumn{1}{|c|}{.   .   .} \\ \hline
        
    \end{tabular}
    \label{tab:example_dialog}
\end{table}

\begin{table}[ht]
\renewcommand*{\arraystretch}{1.2}
    \caption{An example of a conversation snippet from the top performing system. The system is capable of maintaining two simultaneous threads of dialog about League of Legends and music, and switch between them in a coherent and engaging manner.}

    \centering
    \begin{tabular}{|m{\linewidth}|}
    \hline
\textbf{User:} Hello\\
\textbf{System:} hello! hows your day going so far?\\
\textbf{User:} What are you up to?\\
\textbf{System:} just finished making coffee! what about you? what do you do?\\
\textbf{User:} i'm playing video games\\
\textbf{System:} oh, nice! what kind? and i'm just relaxing here and listening to some classical music.\\
\textbf{User:} i'm playing league of legends\\
\textbf{System:} nice! what rank are you? i haven't played in forever.\\
\textbf{User:} who is your favorite classical musician?\\
\textbf{System:} probably beethoven. i like classical music.\\
\textbf{User:} i'm plat 2\\
\textbf{System:} nice! i'd imagine you are better than that!\\
\textbf{User:} can you play an instrument?\\
\textbf{System:} i can play the piano. do you play any instruments?\\
\textbf{User:} thanks\\
\textbf{System:} i appreciate it! so what is it like being plat 2?\\
\textbf{User:} i can play the clarinet, but not very well haha\\
\textbf{System:} oh nice! i played clarinet in high school.\\
\textbf{User:} the games are difficult but rewarding\\
\textbf{System:} yes they are! i played the trumpet in college.\\
\multicolumn{1}{|c|}{.   .   .} \\ \hline
        
    \end{tabular}
    \label{tab:example_dialog2}
\end{table}

While many of the submissions in Sub-task 1 perform similarly, the scores in Table \ref{tab:track3_task2} for Sub-task 2 are significantly more varied. This means that interactive evaluation more exhaustively tests the capabilities of systems and is therefore more indicative of a system's capabilities. This observation has been shown by prior work \cite{mehri2020unsupervised}, when analyzing dialogs from Meena \cite{adiwardana2020towards}.

Tables \ref{tab:example_dialog} and \ref{tab:example_dialog2} show sample dialogs with the top performing system. In both dialogs, we observe that the system produced very relevant and engaging responses. Furthermore, the users appear to be engaged in the interaction, which again highlights the importance of evaluating with real users. In Table \ref{tab:example_dialog2} we see the system maintain two simultaneous threads of dialog, about League of Legends and music. It shifts between them in a natural and engaging manner.

\subsection{Discussion}

\subsubsection{Sub-task 2 Evaluation Metrics:} FED \cite{mehri2020unsupervised}, which is an \textit{unsupervised} evaluation metric for interactive dialog is shown to be a moderate predictor of the final ranking with a system-level Spearman correlation of 0.49 ($p = 0.13$), though it correctly predicts the top two systems. There is still significant room for improvement for the difficult problem of automatic evaluation metrics for interactive settings, where there is no ground-truth response and the domain is unrestricted. 

We also note that the average number of turns for a particular system is a strong indicator of its quality here (Spearman: 0.94, $p < 0.01$). Real users are more inclined to interact with a better system, making it an important metric for assessing systems in interactive settings \cite{ram2018conversational}. This observation brings more evidence to the argument that evaluations should be carried out with real users, They interact with a system of their own volition and terminate the dialog when they are no longer engaged.

\subsubsection{Open-Domain Dialog Systems}

The best performing systems in both sub-tasks relied heavily on pre-trained language models, signifying that large-scale pre-training is vital for handling unconstrained interactions with real users. Furthermore, all of the top 3 models used an evaluation model to re-rank responses and to filter out irrelevant or incoherent ones. This suggests that while pre-trained models are surprisingly effective, the use of a more sophisticated pipeline (e.g., evaluation model, dialog planning model) improves the robustness of a system and results in better interactions. 

\subsubsection{Sub-task 2 Interactive Evaluation Paradigm:}

The \textit{Interactive Evaluation of Dialog track} demonstrates both the feasibility and the importance of evaluating dialog systems in interactive settings with real users. We show that with an advertising budget of \$2500, we collect more than 4000 dialogs on DialPort (2960 dialogs with at least 4 turns or 8 utterances), thus the cost was less than \$1.00 per usable dialog. The DialPort platform, through funding from the National Science Foundation, is able to provide interactive evaluation as a service free of charge to any dialog researchers. As of early 2023, DialPort will be managed by the linguistic data consortium\footnote{\url{https://www.ldc.upenn.edu/}}. 

Furthermore, interactive evaluation poses a unique set of challenges for dialog systems. The results of interactive evaluations are more varied (Table \ref{tab:track3_task2}), suggesting that back-and-forth interactions with real users are challenging to dialog systems and that interactive evaluation better reflects a system's capabilities. Response generation on static datasets neglects several valuable properties of dialog systems, including consistency, topic depth, adaptation, error recovery and user-centric development. 

It is difficult to maintain consistency when evaluating in interactive settings, as there is no way of ensuring that different systems are challenged to the same extent. However, as shown by the Alexa Prize \citep{ram2018conversational}, this problem can be mitigated by collecting enough dialogs such that the average complexity is approximately equal for all systems. In addition, for consistency we ran interactive evaluation for all the systems simultaneously to remove temporal variation.

The results here especially validate the importance of real users, a defining aspect of the DialPort platform. Since users interact with systems out of some perceived interest, they have longer interactions with better systems making average dialog length a strong indicator of system quality.

\section{Acknowledgments}

This work is funded by National Science Foundation grant CNS-1512973. The opinions
expressed in this paper do not necessarily reflect
those of the National Science Foundation.

\section{Conclusion}

This paper describes the \textit{Interactive Evaluation of Dialog track} at the 9th Dialog System Technology Challenge which had the goal of challenging participants to extend dialog models to interactive settings with real users. For Sub-task 1, there were 33 submissions, which reported strong results for static evaluation on the Topical-Chat corpus. For Sub-task 2,  dialog models were evaluated on DialPort with users recruited through Facebook Advertising. Participants developed novel models for both sub-tasks, including approaches for generating more relevant and diverse responses and having more coherent dialogs with users. This challenge demonstrates both the feasibility and value of interactive evaluation. Automatic metrics such as USR and FED were found to correlate moderately with human judgements, and  conversation length is found to be a strong predictor of system quality when assessing with real users. 

\bibliography{aaai.bib}

\end{document}